\pdfoutput=1
\documentclass[11pt]{article}
\usepackage[hyphens]{url}

\usepackage[]{EMNLP2022}

\usepackage{times}
\usepackage{latexsym}
\usepackage{todonotes}
\usepackage[T1]{fontenc}

\usepackage[utf8]{inputenc}

\usepackage{microtype}

\usepackage{inconsolata}

\usepackage{booktabs}
\usepackage{subcaption}
\usepackage{float}
\usepackage[inline]{enumitem}
\usepackage{dsfont}
\usepackage{amsmath}
\usepackage{mathtools}
\newcommand{\defeq}{\vcentcolon=}
\usepackage{siunitx}

\usepackage{amsfonts}
\usepackage{multirow}
\usepackage{graphicx}
\usepackage{physics}
\usepackage{xfrac}
\usepackage{breqn}
\usepackage{enumitem}
\graphicspath{ {./images/} }

\DeclareMathOperator*{\argmax}{arg\,max}

\DeclareMathOperator*{\EntCE}{EntCE}
\DeclareMathOperator*{\RankCS}{RankCS}
\DeclareMathOperator*{\DistCE}{DistCE}

\newcommand{\ie}{\emph{i.e., }}
\newcommand{\eg}{\emph{e.g., }}

\title{Stop Measuring Calibration When Humans Disagree}
\author{Joris Baan${^{1}}$, Wilker Aziz${^1}$, Barbara Plank${^{2,3,4}}$, Raquel Fern{\'a}ndez${^1}$ \\
        ${^1}$University of Amsterdam, ${^2}$IT University of Copenhagen,
        ${^3}$MCML Munich,
        ${^4}$LMU Munich
        \\
        \texttt{\{j.s.baan,w.aziz,raquel.fernandez\}@uva.nl},  \texttt{b.plank@lmu.de}}
\begin{document}
\maketitle

\begin{abstract}
Calibration is a popular framework to evaluate whether a classifier \textit{knows when it does not know}---\ie its predictive probabilities are a good indication of how likely a prediction is to be correct. Correctness is commonly estimated against the human majority class. Recently, calibration to human majority has been measured on tasks where humans inherently disagree about which class applies. We show that measuring calibration to human majority given inherent disagreements is theoretically problematic, demonstrate this empirically on the ChaosNLI dataset, and  derive several instance-level measures of calibration that capture key statistical properties of human judgements---class frequency, ranking and entropy.\footnote{Code available at \url{https://github.com/jsbaan/calibration-on-disagreement-data}.}
\end{abstract}

\section{Introduction} \label{section:introduction}

Neural text classifiers are becoming more powerful but increasingly difficult to interpret~\cite{rogers2020primer}. In response, the demand for transparency and trust in their predictions is growing~\cite{yin2019understanding,bansal2019beyond,bianchi-hovy-2021-gap}. One step towards understanding when to trust predictions is to evaluate whether models \textit{know when they do not know}---\ie whether predictive probabilities are a good indication of how likely a prediction is to be correct---known as \textit{calibration}. This is crucial in user-facing and high-stake applications. 

Calibration, and particularly the Expected Calibration Error~\cite[ECE;][]{naein-obtaining-2015, guo-on-calibration-2017}, is widely studied in Machine Learning and Computer Vision~\cite{mena-survey-2022}, and is gaining increased attention in Natural Language Processing~\cite[NLP;][]{desai-durrett-2020-calibration, kong-etal-2020-calibrated, jiang-how-can-2021, dan-roth-2021-effects-transformer}. 

An important implicit \textit{assumption} in the widely used definition of perfect calibration proposed by \citet{guo-on-calibration-2017} is that predictions are \textit{either right or wrong}---in other words, that the true class distribution, \ie human judgement distribution, is deterministic (one-hot). However, for many problems, while categories exist, their boundaries are fluid: there exists \textit{inherent disagreement} about labels. This means that gold labels are at best an idealization---as irreconcilable disagreement is abundant~\cite{plank2014linguistically,aroyo&welty:AIMagazine15,jamison-gurevych-2015-noise,palomaki2018case,pavlick-kwiatkowski-2019-inherent}. Evidence for this can be found in various tasks, including those which involve linguistic and subjective judgements~\cite{Akhtar:HCOMP20, basile2021we}. 
%
Surprisingly, however, while limitations of calibration are studied (\S\ref{section:background}), this fundamental assumption is ignored. 

In this work, we show that popular calibration metrics---such as ECE---are not applicable to data with inherent human disagreement (\S\ref{section:problems}). We propose an alternative, instance-level notion of calibration based on human uncertainty, and operationalize it with several measures that capture key statistics of the human judgement distribution other than matching the majority vote (\S\ref{section:human-uncertainty}). Finally, we verify our theoretical claims with a case study on the ChaosNLI dataset, and investigate temperature scaling---a popular post-hoc calibration method---through the lens of human uncertainty (\S\ref{section:experiments}).


\section{Background\label{section:background}}

\paragraph{Data} We have data $\mathcal D = \{(x_n, y_n)\}_{n=1}^{N}$ where $x_n$ is an instance (\ie text or texts) and $y_{n} \!\in\! [C]$ is a category.  
For any instance $X\!=\!x$,\footnote{We use capital letters for random variables (\eg $X$) and lowercase letters for their assignments (\eg $X=x$), $[C]$ is short for $\{1, \ldots, C\}$, $[a=b]$ is the Iverson bracket (\ie it evaluates to $1$ if the predicate $a=b$ is True, 0 otherwise), and $\Delta_{C-1} \subset \mathbb R^C$ is the simplex (\ie set of $C$-dimensional vectors whose coordinates are positive and sum to 1).} we assume that human annotators draw their labels independently from the same Categorical distribution
with class probabilities $\pmb \pi(x) \!\in\! \Delta_{C-1}$. 
That is, the probability $\Pr(Y=c|X=x)$ that a human should label $x$ an instance of $c \in [C]$ is $\pi_c(x)$. 
 For observed $x$, an estimate of $\pmb \pi(x)$ can be obtained via maximum likelihood estimation (MLE). This estimate $\bar{\pmb \pi}(x)$ is the vector whose coordinate $\bar{\pi}_{c}(x) = \frac{\sum_{n=1}^N [x_n = x][y_n = c]}{\sum_{n=1}^N [x_n = x]}$ is the relative frequency with which $x$ is labeled as $c \in [C]$. 
 Oftentimes, $\pmb \pi(x)$ is assumed to be one-hot (\ie the task is unambiguous), in such cases, a single human judgement per instance is sufficient for an exact estimate. 

\paragraph{Classification}
A probabilistic classifier approximates $\pmb \pi(x)$ with a trained parametric function \citep[\eg BERT;][]{devlin-etal-2019-bert} that maps an input $x$ to a vector $\mathbf f(x)$ of class probabilities. 
After training, and given an instance $x$, we typically map the model's output $\mathbf f(x)$ to a single decision $\hat y \in [C]$. More often than not, this is the mode of the model distribution: $\hat y = \argmax_{c}~f_c(x)$. 
The correctness of $\hat y$ is assessed against the observed human `gold standard' decision $y^\star = \argmax_{c}~\bar\pi_c(x)$. 


\paragraph{Calibration}
A classifier is \emph{multi-class calibrated} \citep{vaicenavicius2019evaluating, kull2019beyond} if, for all instances mapped to the same vector $\mathbf q$, the relative frequency with which $c$ is correct (assessed against the gold standard) is  $q_c$ for every $c$: 
\begin{align}
    \label{eq:multi-class-calibration}
    \Pr(Y^\star = c \mid \mathbf f(X)\!=\!\mathbf q) = q_c \quad \forall c \in [C]
\end{align}
Consider a problem with three classes. A model is multi-class calibrated if, for all instances mapped to the same vector, \eg $(0.90, 0.07, 0.03)^\top$, predicting the first class would result in a correct decision for 90\% of these instances, the second class for 7\%, and the third class for 3\%. 
Estimation of the left-hand side (LHS) of Eq(\ref{eq:multi-class-calibration}) by counting is difficult as it requires observing multiple instances mapped to the same probability vector. 

A weaker notion of calibration \cite[popular in NLP;][]{desai-durrett-2020-calibration, jiang-how-can-2021} is \textit{confidence calibration}~\citep{guo-on-calibration-2017}:\footnote{The maximum class probability assigned to a random text $X$ is itself a random variable, we denote it by $\max(\mathbf f(X)) \defeq \max_{c \in [C]}~f_c(X)$. As for any other random variable,  $\max(\mathbf f(X))=p$ identifies the set of texts for which the maximum probability predicted by the model is exactly $p$.}
\begin{align}
    \label{eq:confidence-calibration}
    \Pr(Y^\star = \hat Y \mid \max(\mathbf f(X)) = p) = p
\end{align}
A model is confidence calibrated if, for all instances mapped to a maximum probability value $p$ (\eg 0.9), the most probable class under the model is correct for 90\% of these instances.

\paragraph{Expected Calibration Error}\label{section:ece} is most often used to measure (confidence) calibration in practice. \citet{naein-obtaining-2015} originally proposed ECE for binary classification 
and \citet{guo-on-calibration-2017} later adapted it to a multi-class setting:
\begin{align}
    \label{eq:ece}
    \mathrm{ECE} &= \sum_{m=1}^M\frac{|B_m|}{N}|\mathrm{acc}(B_m)-\mathrm{conf}(B_m)|
\end{align}
ECE estimates the confidence calibration error---absolute difference between the LHS and the RHS of Eq(\ref{eq:confidence-calibration})---in expectation by discretizing the probability of the model decision into a fixed number $M$ of intervals (or bins). Each prediction vector $\mathbf f(x)$ is assigned to a bin $B_m$ based on its highest probability $\max_c~f_c(x)$. 
The ECE is the weighted average of the difference between the average confidence and accuracy per bin. To obtain zero calibration error, if 90 out of 100 instances that received a highest probability between 0.8 and 1.0 are correctly classified, the average confidence on those 100 instances must be 0.9.

Several recent studies identify and address problems with ECE---mostly with its binning scheme and implicit decision rule \cite[\eg][]{kumar-trainable-2018, nixon_measuring-2019,widmann2019calibration, gupta2021calibration, si2022revisiting}. Instead, in this work we identify a fundamental problem in the definition of perfect calibration when applying it to setups where there exists no real gold label. 

\section{Calibration \& Disagreement Pathology} \label{section:problems}

It is common practice to handle human disagreement with majority voting or other aggregation methods~\cite{dawid1979maximum,artstein2008inter,paun2022statistical}. Aggregate (gold) labels are then used to evaluate a classifier's accuracy. We now illustrate the problem this poses when measuring calibration.

\begin{description}[leftmargin=10pt]
\item[Desideratum:] \textit{Any classifier $\mathbf g$ that, given an instance $x$, predicts the human judgement distribution $\mathbf g(x)=\pmb \pi(x)$ should be perfectly calibrated.}
\end{description}
Consider the oracle classifier that has access to the MLE $\bar{\pmb \pi}(x)$ of $\pmb\pi(x)$ for any instance $x$ in a validation set. 
For each $x$, the oracle is able to predict the human labeling uncertainty $\pmb\pi(x)$. 
This estimate is unbiased and becomes more precise the more judgements we have access to.
By definition, when human majority voting is used, this classifier achieves perfect accuracy---its highest confidence prediction always matches the gold standard. 
However, according to ECE, the oracle classifier is miscalibrated (this is true for other definitions of calibration to accuracy, including multi-class and classwise). Recall that the calibration error is the absolute difference between accuracy and average confidence per bin. The accuracy of the oracle classifier is always 1. On data where humans disagree, the average confidence will be lower than 1.

This mismatch results in high calibration error (as demonstrated in \S\ref{section:experiments}) and exposes a problem with using ECE to measure calibration on disagreement data. An important takeaway is that, even if we can train a classifier that perfectly models the human judgement distribution, this classifier would still be severely miscalibrated. To achieve perfect calibration, its probabilities must drift towards an unfaithful representation of human confidence. Therefore, we argue that human majority accuracy is a bad estimate of correctness to calibrate against.

\begin{figure*}[ht]
     \centering
     \begin{subfigure}[b]{0.24\textwidth}
         \centering
         \includegraphics[width=\textwidth]{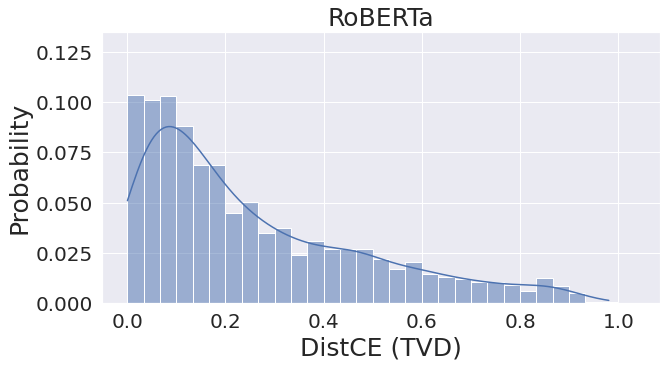}
         \caption{$\DistCE$: Vanilla}
         \label{fig:main-figurea}
     \end{subfigure}
     \hfill
     \begin{subfigure}[b]{0.24\textwidth}
         \centering
        \includegraphics[width=\textwidth]{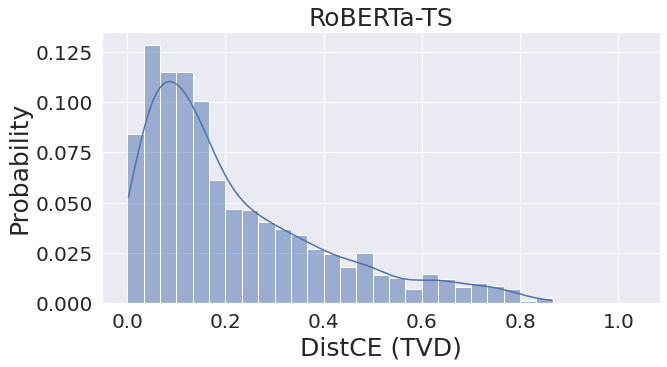}
         \caption{$\DistCE$: Temp Scaling}
         \label{fig:main-figureb}
     \end{subfigure}
     \hfill
     \begin{subfigure}[b]{0.24\textwidth}
         \centering
         \includegraphics[width=\textwidth]{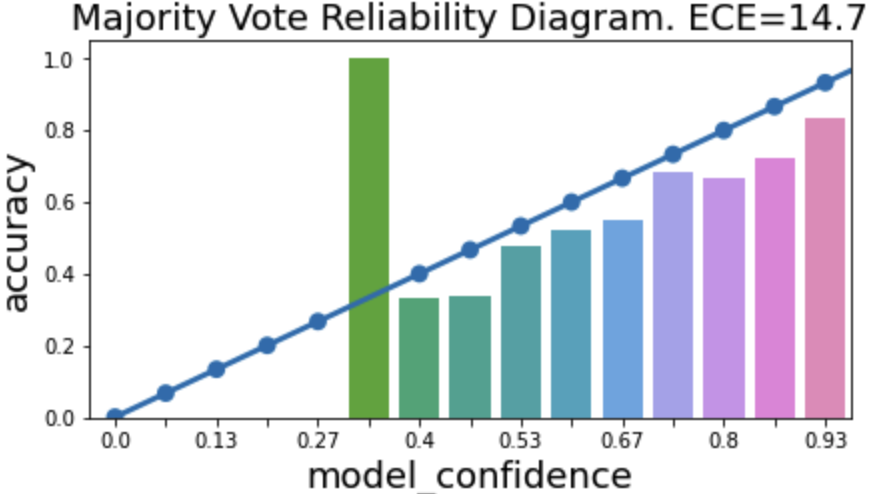}
         \caption{ECE: Vanilla}
         \label{fig:main-figurec}
     \end{subfigure}
     \begin{subfigure}[b]{0.24\textwidth}
         \centering
         \includegraphics[width=\textwidth]{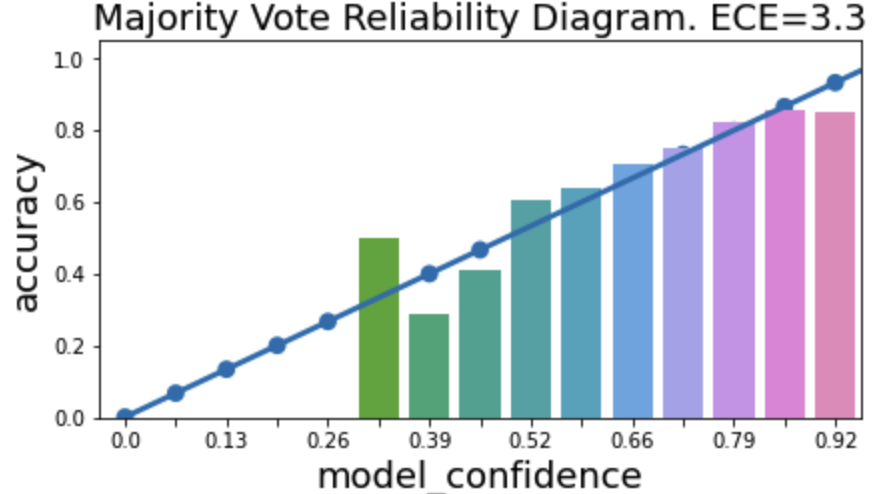}
         \caption{ECE: Temp Scaling}
         \label{fig:main-figured}
     \end{subfigure}
    \caption{Left: Distribution over instance-level calibration errors ($\DistCE$). While temperature scaling (TS) causes fewer severely miscalibrated instances---illustrated by the right tail of the distribution retracting from (a) to (b)---there are \textit{also} fewer instances that are perfectly calibrated, see drop of first bar in (b). Right: Reliability Diagrams indicate that TS improves ECE enormously, because the bars in (d) move towards the diagonal.}
    \label{fig:main-figure}
\end{figure*}

\section{Calibration to Human Uncertainty} \label{section:human-uncertainty}

To get a faithful probabilistic classifier, we expect it to predict the uncertainty the human population exhibits on any given $x$. 
Notions of calibration to accuracy (\eg multi-class, classwise, confidence) are defined \emph{marginally} (\ie for instances grouped by a property of model predictions such as  their probability). Instead, we argue for a direct assessment of calibration at the instance level. Given $x$, perfect calibration to human uncertainty requires: 
\begin{equation}\label{eq:noveldef}
\Pr(Y=c \mid X=x) = f_c(x) \quad \forall c \in [C] 
\end{equation}
This is our desideratum of \S\ref{section:problems} re-expressed for a practical classifier $\mathbf f(\cdot)$.
In words, a model is calibrated for $x$ if it predicts probability $f_c(x)$ equal to the probability $\Pr(Y=c \mid X=x)$  with which humans label $x$ as $c$. 
With multiple human judgements (whether or not they disagree), the LHS can be estimated by $\bar{\pmb\pi}(x)$---the relative frequency of the observed labels. 
Assessing the degree to which Eq(\ref{eq:noveldef}) holds in expectation across instances gives us a tool to criticize classifiers in terms of their overall calibration to human uncertainty in a given task. This is appealing because we can assess the trustworthiness both globally (overall calibration) as well as on individual predictions.

\paragraph{Distance Measures} \label{section:distance}
To operationalize our notion of human calibration, we propose three distance measures---each capturing a different key statistic. 
First, Human Entropy Calibration Error:
\begin{equation}
    \EntCE(x) = H(\mathbf f(x)) - H(\bar{\pmb \pi}(x)) 
\end{equation}
This captures the alignment between disagreement among humans and a model's \textit{indecisiveness}. It is sensitive to average confusion, but not to class ranking.
Second, Human Ranking Calibration Score: 
\begin{multline}
    \RankCS = \frac{1}{N} \sum_{n=1}^N [\mathrm{argsort}(\mathbf f(x_n)) = \\[-10pt] \mathrm{argsort}(\bar{\pmb\pi}(x_n))]
\end{multline}
$\RankCS$ is a global measure that can be viewed as a stricter alternative to majority vote accuracy. It is sensitive to class ranking but not to magnitude of probability, complementing entropy calibration.\footnote{We leave aside specifying a method for handling cases where two or more classes obtain an equal number of votes, which makes multiple rankings correct.} Third, Human Distribution Calibration Error---the strictest and most informative measure of the three:
\begin{equation}
    \DistCE(x) = \mathrm{TVD}(\mathbf f(x), \bar{\pmb \pi}(x)) 
\end{equation}
We opt for the popular total variation distance (TVD) between the predictive distribution and the human judgement distribution.\footnote{$\operatorname{TVD}(\mathbf q, \mathbf p) = \sfrac{1}{2} \norm{\mathbf q - \mathbf p}_1 $ is convenient for various reasons: it is a metric (hence, symmetric) and defined for all discrete distributions (including sparse distributions); being expressed directly in (absolute difference in) probability, it is dimensionless (or unitless) and bounded between 0 and 1; it quantifies the maximum discrepancy in probability across the event space. For a more technical discussion, see Appendix \ref{ap:TVD}.}
One could compare other statistics of interest, and we encourage the community to do so. For example, a more fine-grained classwise analysis (see Appendix~\ref{ap:experiments}).%

\paragraph{Advantages}
First, unlike ECE, human calibration naturally handles disagreement data---in fact, it requires multiple annotations to reliably estimate the human judgement distribution. 
Second, $\DistCE$ and $\EntCE$ measure the calibration of \textit{individual} predictions, which is a powerful tool to aid decision making.
Crucially, this avoids the need for a binning scheme, often criticized in ECE \citep{nixon_measuring-2019, gupta2021calibration}.
Third, $\DistCE$ ensures full multi-class calibration: there is no implicit decision rule and a classifier's underlying statistical model is directly evaluated on its ability to match the entire human judgement distribution.
Fourth, unlike ECE and its variants, $\DistCE$ is a proper scoring rule, which comes with a range of desirable properties \citep{gneiting2007strictly}.

\paragraph{Related Work}
Several recent studies on \textit{soft evaluation} evaluate or optimize for a quantity similar to our $\DistCE$.
However, we are the first to propose a general notion of calibration in disagreement settings.
\citet{nie-etal-2020-learn} introduce ChaosNLI (see \S\ref{section:experiments}) and report the Kullback–Leibler (KL) and Jensen–Shannon (JS) divergences from estimates of the human judgement distributions. 
Follow-up work explores decreasing this divergence. 
For example, by fine-tuning on them directly~\citep{meissner-etal-2021-embracing, zhang-etal-2021-learning-different}; using Bayesian-inspired 
methods~\citep{zhou-etal-2022-distributed}; methods popularized to bring down ECE, such as temperature scaling or label smoothing~\citep{zhang-etal-2021-learning-different, wang-etal-2022-capture}; or framing the task as regression \citep{chen-etal-2020-uncertain}. 

We show that the human majority class is not a meaningful statistic to calibrate against in disagreement settings. In \S\ref{section:experiments}, we empirically demonstrate that \textit{human calibration} is more faithful, and a useful tool to gain insights into calibration errors.

\section{Case Study} \label{section:experiments}
\subsection{Experimental Setup}
\paragraph{Dataset}

We use the ChaosNLI dataset \cite{nie-etal-2020-learn} as case study. It contains English natural language inference instances selected from the development sets of SNLI \citep{bowman-etal-2015-snli}, MNLI \citep{williams-mnli-2018} and AbductiveNLI \citep{bhagavatula2020abductive} for having a borderline annotator agreement, \ie at most 3 out of 5 human votes for the same class. ChaosNLI collects an additional 100 independent annotations for each of the roughly 1,500 instances per dataset, resulting in $T=100$ human votes distributed over $C=3$ classes per premise-hypothesis pair for a total of $N=4,645$ instances. The dataset was collected very carefully and with strict annotation guidelines. This ensures that disagreement cannot easily be discarded as noise \citep{pavlick-kwiatkowski-2019-inherent, nie-etal-2020-learn}. The task description and examples can be found in Appendix~\ref{ap:examples}.

\paragraph{Method}
We fine-tune RoBERTa \citep{liu2019roberta} on SNLI following the standard procedure described by \citet{desai-durrett-2020-calibration}. We evaluate on the ChaosNLI-SNLI split.
To investigate the value of human calibration, we inspect \citet{wang-etal-2022-capture}'s claim that ECE is a good alternative to measuring divergence to the human judgement distribution---and that temperature scaling~\cite[TS;][]{guo-on-calibration-2017} is a suitable calibration method to do so. We discuss temperature scaling and how we choose a temperature in Appendix \ref{ap:ts}.%

\subsection{Results}
Table \ref{tab:acc-ece} shows accuracy, ECE,\footnote{We observe a similar trend for classwise-ECE, a popular improvement to standard ECE, in Appendix~\ref{ap:classwise-ece}.} $\RankCS$, and summary statistics of the instance-level $\EntCE$ and $\DistCE$ metrics for RoBERTa, temperature scaled RoBERTa-TS and the oracle classifier.

\paragraph{Oracle is miscalibrated} Indeed, the oracle classifier is severely miscalibrated according to ECE---even more so than RoBERTa (0.25 vs 0.14), demonstrating the problem we highlight in \S\ref{section:problems}. Instead, on all our human calibration metrics, the oracle is perfectly calibrated.

\paragraph{Inspecting Error Distributions}
Applying TS to RoBERTa results in a sharp decrease in ECE (from 0.14 to 0.03). The reliability diagrams\footnote{A reliability diagram \cite{degroot-comparison-1983, naein-obtaining-2015} visualizes ECE by plotting accuracy against confidence (intervals). As bars approach the diagonal, ECE converges to 0.} in Figures~\ref{fig:main-figurec} and \ref{fig:main-figured} confirm this, suggesting that TS successfully calibrates probability values.
\begin{table}[h!]
\centering
\resizebox{\columnwidth}{!}{
\begin{tabular}{@{}l@{\hspace{-15pt}} S[table-number-alignment = center]
@{\hspace{-10pt}} S[table-number-alignment = center]
S[table-number-alignment = center]
@{}}
\toprule
    & \multicolumn{1}{@{\hspace*{20pt}}c}{RoBERTa} & \multicolumn{1}{r}{RoBERTa-TS} & \multicolumn{1}{c@{}}{Oracle} \\ \midrule
Acc $\uparrow$ & 0.74 \textpm 0.01 & 0.74 \textpm 0.01 & 1.00\\
ECE $\downarrow$ & 0.14 \textpm 0.01 & 0.03 \textpm 0.01 & 0.25 \\ \midrule
$\RankCS \uparrow$ & 0.62 \textpm 0.01 & 0.62 \textpm 0.01 & 1.00\\ 
$\mathbb E[|\EntCE|] \downarrow$ & 0.30 \textpm 0.02 & 0.21 \textpm 0.02 & 0.00\\
$\mathbb E[\DistCE] \downarrow$ & 0.26 \textpm 0.00 & 0.22 \textpm 0.00 & 0.00\\ 
\bottomrule
\end{tabular}
}
\caption{\label{tab:acc-ece} Accuracy \& ECE versus our $\RankCS$ and summary (mean) of instance-level $|\EntCE|$ \& $\DistCE$ on ChaosNLI-SNLI. Results shown over 3 random seeds.} 
\end{table}

However, TS only causes a very small change in mean $\DistCE$ (from 0.26 to 0.22). Though the practical significance of this shift might not be immediately obvious (that is also true for other metrics, such as ECE), human calibration allows us to inspect how errors are distributed \textit{across instances}. This is an important tool to gain more insight into the effects of a method such as TS on calibration. 

The global $\DistCE$ error distributions in Figure~\ref{fig:main-figurea} and \ref{fig:main-figureb} reveal that perfectly-calibrated instances are sacrificed to reduce poorly-calibrated instances. This corroborates our intuition that TS artificially compresses the predicted probability range, which, arguably is not desirable. 
For more extensive and fine-grained analyses, including out-of-distribution evaluation on the ChaosNLI-MNLI set, see Appendix \ref{ap:experiments}.

\paragraph{How Good is Good?}
A naturally arising question is how \textit{good} the shape of a $\DistCE$ error distribution is. To answer this, we need a target: what does the error distribution of a \textit{``as-good-as-it-realistically-gets''}-classifier look like?

\begin{figure}[t]
     \centering
     \includegraphics[width=\columnwidth]{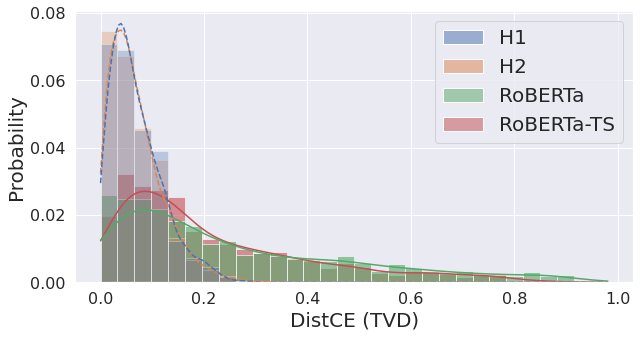}
    \caption{$\DistCE$ error distributions for two \textit{``as-good-as-it-realistically-gets''}-classifiers and two RoBERTas (vanilla and TS). There are big differences between but not within groups.}
    \label{fig:scale}
\end{figure}

To approximate such a classifier, for each premise-hypothesis pair, we sub-sample 20 votes and use them to construct a higher-variance MLE of the underlying human judgement distribution $\pmb \pi(x)$. We construct two such classifiers (H1 and H2) and plot their $\DistCE$ distribution alongside RoBERTa and RoBERTa-TS in Figure~\ref{fig:scale}. We evaluate the classifiers against the 100 available annotations.

As expected, we barely observe any difference between the two sub-sampled human classifiers. However, there is a massive difference between the RoBERTa-based and human-based classifiers---with RoBERTas stretching to much higher instance level calibration errors (x-axis) than humans---thereby providing a sense of \textit{scale}. To quantify these differences, we compute KL-divergences between $\DistCE$ error distributions. We opt for KL-divergence because it is asymmetric; weighting differences in bins that are not probable under the human model less than those that are. We also report TVD. 

Table~\ref{table:scale} shows that KL divergences from RoBERTa or RoBERTa-TS to an ideal classifier's error distribution are 150-170x bigger (0.611 and 0.688) compared to the control group (one ideal classifier to another; 0.004). Even though RoBERTa-TS shows slightly reduced KL-divergence compared to the vanilla model, it is nowhere near an ideal classifier, and it is unclear whether the observed reduction in KL translates to a meaningful or practical difference, \ie something a practitioner would care about.

\begin{table}[h]
    \centering
     \resizebox{\columnwidth}{!}{
    \begin{tabular}{@{}lS[table-number-alignment = center]S[table-number-alignment = center]@{}}
        \toprule
        $\DistCE$ error distribution & \multicolumn{1}{c@{}}{$\mathrm{KL}(\text{H1}, \cdot)$} & \multicolumn{1}{c@{}}{$\mathrm{TVD}(\text{H1}, \cdot)$} \\ \midrule
        H2         & 0.004 & 0.022\\ 
        RoBERTa    & 0.688 & 0.500  \\
        RoBERTa-TS & 0.611 & 0.454  \\ \bottomrule
    \end{tabular}
     }
    \caption{Quantifying divergence between $\DistCE$ error distributions for human sub-populations (H1 and H2) and  neural classifiers (RoBERTa and RoBERTa-TS).}
    \label{table:scale}
\end{table}


\section{Conclusion}
We demonstrate a fundamental problem with measuring calibration to the human majority vote in settings with inherent disagreement in human labels. 
We propose an alternative, instance-level notion based on the full human judgement distribution and operationalize this notion with three metrics. 
We study temperature scaling RoBERTa on the ChaosNLI dataset using these metrics and conclude that they---and crucially, the ability to inspect them in distribution---provide a more robust and faithful lens to analyze classifier calibration in disagreement settings.

Human uncertainty can be used to evaluate many other calibration techniques---we only performed a preliminary analysis for temperature scaling---and we encourage the community to look into those, in addition to exploring other datasets with inherent disagreements.



\section*{Limitations}
A reliable estimate of the human judgement distribution is an important requirement for human calibration. For the ChaosNLI dataset, the reliability is endorsed by the large number of annotations per instance and strict quality control \cite{nie-etal-2020-learn}. Most datasets do not provide this. 
We believe, however, that the advantages of collecting additional annotation outweigh its cost, since, without it, datasets are likely to under-represent human disagreement. 
We therefore advocate future datasets to include multiple annotations per instance (at least for a small test set), as recently also advocated by, \eg\citet{prabhakaran-etal-2021-releasing}, and a better understanding of \textit{how many} annotations are required for good estimates of human uncertainty. 
An important challenge is to distinguish inherent disagreement from noise, for example due to spammers, which negatively affects data quality \cite{Raykar2012eliminating, aroyo2019crowdsourcing, klie-2022-annotation}.

Another limitation is that uncertainty estimates from one population cannot be said to be universally correct. The notion that, given an instance, one unique human distribution governs all annotators is a simplification. Even a single collection of votes from one experiment might contain sub-populations, in which case the marginal distribution is not representative of individual components.


\section*{Acknowledgements}
We thank the anonymous reviewers and meta-reviewer for their invaluable feedback. This project has received funding by the ELLIS Amsterdam Unit. BP is supported by the Independent Research Fund Denmark (DFF) Sapere Aude grant 9063-00077B and ERC Consolidator Grant 101043235. RF is supported by ERC Consolidator Grant 819455.

\bibliography{anthology,custom}
\bibliographystyle{acl_natbib}

\appendix


\section*{Appendix}

\section{ChaosNLI} \label{ap:examples}

\begin{figure}[ht]
    \centering
    \includegraphics[width=\columnwidth]{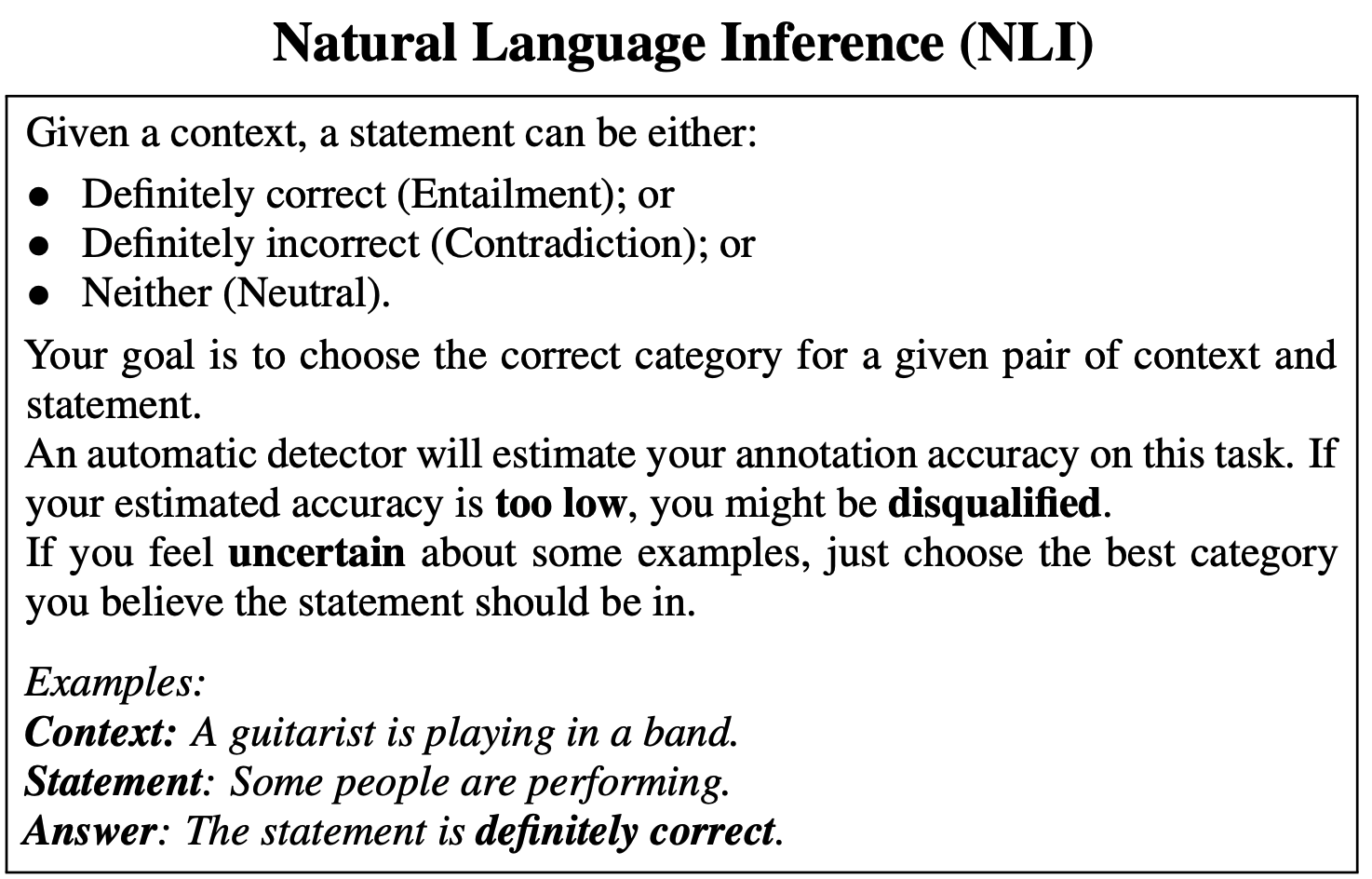}
    \caption{The task description for natural language inference by \citet{nie-etal-2020-learn} used to collect ChaosNLI-SNLI and ChaosNLI-MNLI.}
    \label{fig:task-description}
\end{figure}

Table~\ref{table:high-agreement} and Table \ref{table:low-agreement} show examples with very low or very high agreement. The task description provided by \citet{nie-etal-2020-learn} is in Figure~\ref{fig:task-description}.\footnote{ChaosNLI can be downloaded at \url{https://www.dropbox.com/s/h4j7dqszmpt2679/chaosNLI_v1.0.zip.}}

\begin{table*}[]
\begin{tabular}{p{0.3\linewidth} | p{0.3\linewidth} | p{0.3\linewidth}}
\toprule
Premise & Hypothesis & Label Distribution \\ \midrule
  \multicolumn{1}{p{0.3\linewidth} |}{A man running a marathon talks to his friend} &
  \multicolumn{1}{p{0.3\linewidth} |}{There is a man running.} &
  \multicolumn{1}{p{0.3\linewidth} |}{Entailment: 100, Neutral: 0, Contradiction: 0} \\ \midrule
  
  \multicolumn{1}{p{0.3\linewidth} |}{A young girl plays with a neon-colored Slinky in a crowd of people on a street lined with flags.} &
  \multicolumn{1}{p{0.3\linewidth} |}{The girl is five years old.} &
  \multicolumn{1}{p{0.3\linewidth} |}{Entailment: 0, Neutral: 100, Contradiction: 0} \\ \midrule
  
  \multicolumn{1}{p{0.3\linewidth} |}{A well built black man stands in the subway, listening to headphones. } &
  \multicolumn{1}{p{0.3\linewidth} |}{A man with headphones on is standing in a subway.} &
  \multicolumn{1}{p{0.3\linewidth} |}{Entailment: 100, Neutral: 0, Contradiction: 0} \\ \midrule
\end{tabular}
\caption{High agreement examples from the ChaosNLI dataset.}
\label{table:high-agreement}
\end{table*}

\begin{table*}[]
\begin{tabular}{p{0.3\linewidth} | p{0.3\linewidth} | p{0.3\linewidth}}
\toprule
Premise & Hypothesis & Label Distribution \\ \midrule
  \multicolumn{1}{p{0.3\linewidth} |}{The important thing is to realize that it's way past time to move it. } &
  \multicolumn{1}{p{0.3\linewidth} |}{It cannot be moved, now or ever.} &
  \multicolumn{1}{p{0.3\linewidth} |}{Entailment: 34, Neutral: 32, Contradiction: 34} \\ \midrule
  
  \multicolumn{1}{p{0.3\linewidth} |}{An elderly woman crafts a design on a loom.} &
  \multicolumn{1}{p{0.3\linewidth} |}{The woman is sewing.} &
  \multicolumn{1}{p{0.3\linewidth} |}{Entailment: 35, Neutral: 31, Contradiction: 34} \\ \midrule
  
  \multicolumn{1}{p{0.3\linewidth} |}{Number 13 kicks a soccer ball towards the goal during children's soccer game.} &
  \multicolumn{1}{p{0.3\linewidth} |}{A player passing the ball in a soccer game.} &
  \multicolumn{1}{p{0.3\linewidth} |}{Entailment: 36, Neutral: 33, Contradiction: 31} \\ \midrule
\end{tabular}
\caption{Low agreement examples from the ChaosNLI dataset.}
\label{table:low-agreement}
\end{table*}

\section{Additional Analyses} \label{ap:experiments}

This section provides additional analyses and figures. 
Each figure shows a comparison between model predictions and human predictions using the measures we propose in \S\ref{section:distance}. We use three different views to compare statistics of human judgements to statistics of model predictions.\footnote{Note that these views do not apply to the $\DistCE$ (row 3 column 1) and reliability diagrams (row 4), which will not change across views.}

The first view shows two \textit{marginal} or dataset-level histograms: one of a human statistic and one of a model statistic, \eg entropy. This view is useful to compare the global distribution over an instance-level statistic between human judgements and model predictions (Figure \ref{figure:calibration-fig-snli-full-temp1.0} and \ref{figure:calibration-fig-snli-full-temp2.0}).

The second view shows one histogram of the \textit{conditional} instance-level error between a model and human statistic, \eg entropy (Figure \ref{figure:calibration-fig-snli-error-temp1.0} and \ref{figure:calibration-fig-snli-error-temp2.0}). This is interesting for diagnosing a classifier's under or over confidence. Instances centered around zero have zero error, instances in the positive range exhibit over-confidence, and instances in the negative range under-confidence.

The third view is similar to the previous, (it is also \textit{conditional}, \eg it compares an instance-level statistic between humans and a model) but shows the \textit{absolute} errors (Figure \ref{figure:calibration-fig-snli-abs_error-temp1.0} and \ref{figure:calibration-fig-snli-abs_error-temp2.0}). This is useful to spot general miscalibration, regardless of the direction (\ie under-confidence vs over-confidence). 

In each figure, the top row shows a histogram of the predicted probability for class 0 (entailment), 1 (neutral) and 2 (contradiction). The second row shows predicted probability for the $k$th highest predicted probability, \ie the first, second and third guess from either the model or human distribution (note that the corresponding classes are not necessarily the same for the model and humans---this row is informative to compare the magnitude of the probability for the same rank). The third row shows the histogram of $\DistCE$ (TVD) from \S\ref{section:experiments} (left) and the histogram for $\EntCE$ (right). The fourth row shows conventional reliability diagrams that visualize ECE, and the number of instances per bin (note that this is not normally shown---though we find it very insightful).

\subsection{Beyond DistCE}\label{ap:beyond}
\begin{figure*}[ht]
\includegraphics[width=0.65\textheight]{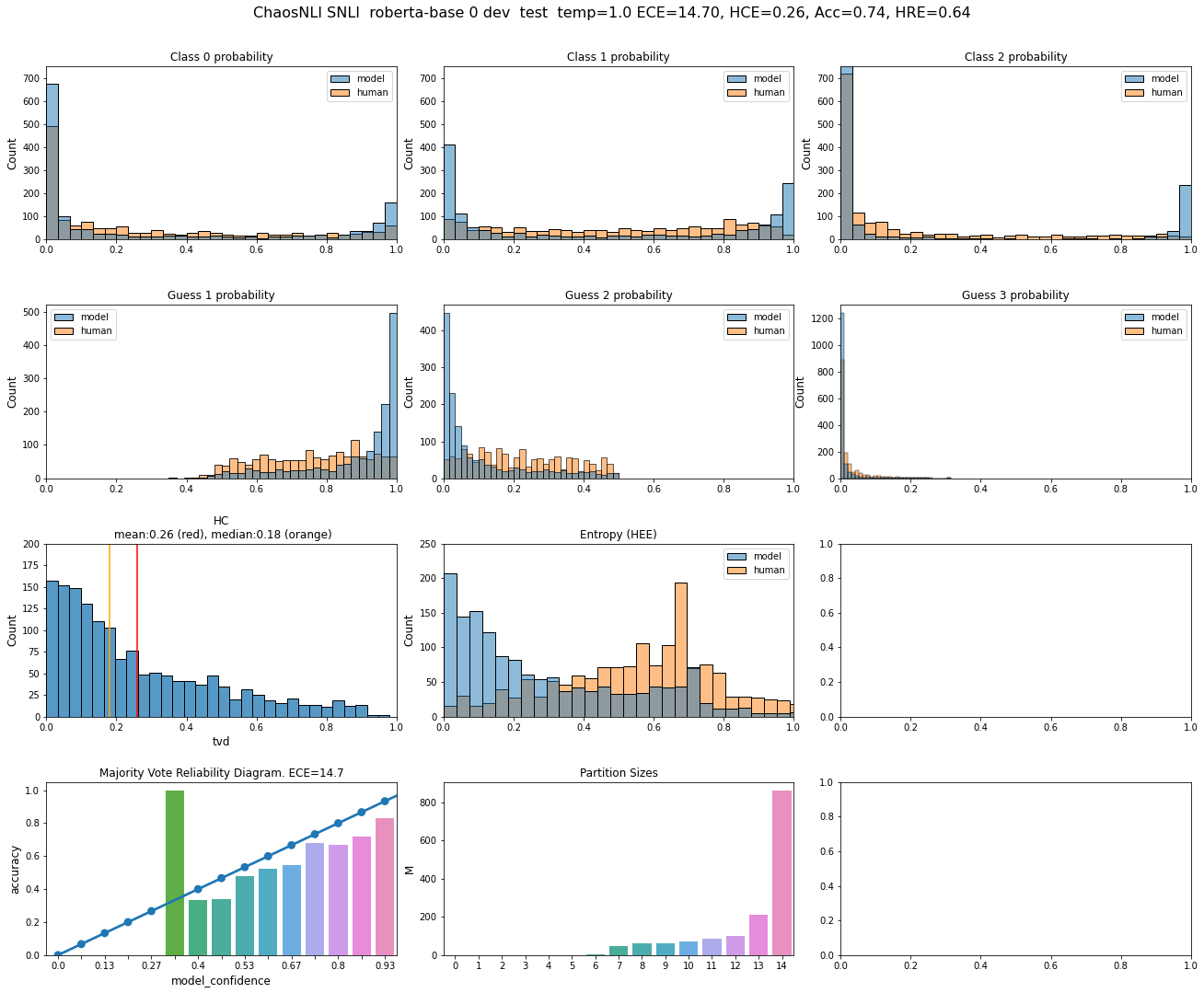}
\caption{RoBERTa-0 on the ChaosNLI-SNLI dev+test set. Several figures comparing human uncertainty to model uncertainty using TVD, confidence, entropy, and reliability diagrams. This figure shows the distribution over instance-based absolute errors between probabilities for each class (top row) or the model vs human $k$th guess (i.e., the highest model probability versus the highest human probability on each instance). See Appendix \ref{ap:experiments} for more information.}
\label{figure:calibration-fig-snli-full-temp1.0}
\end{figure*}

\begin{figure*}[ht]
\includegraphics[width=0.65\textheight]{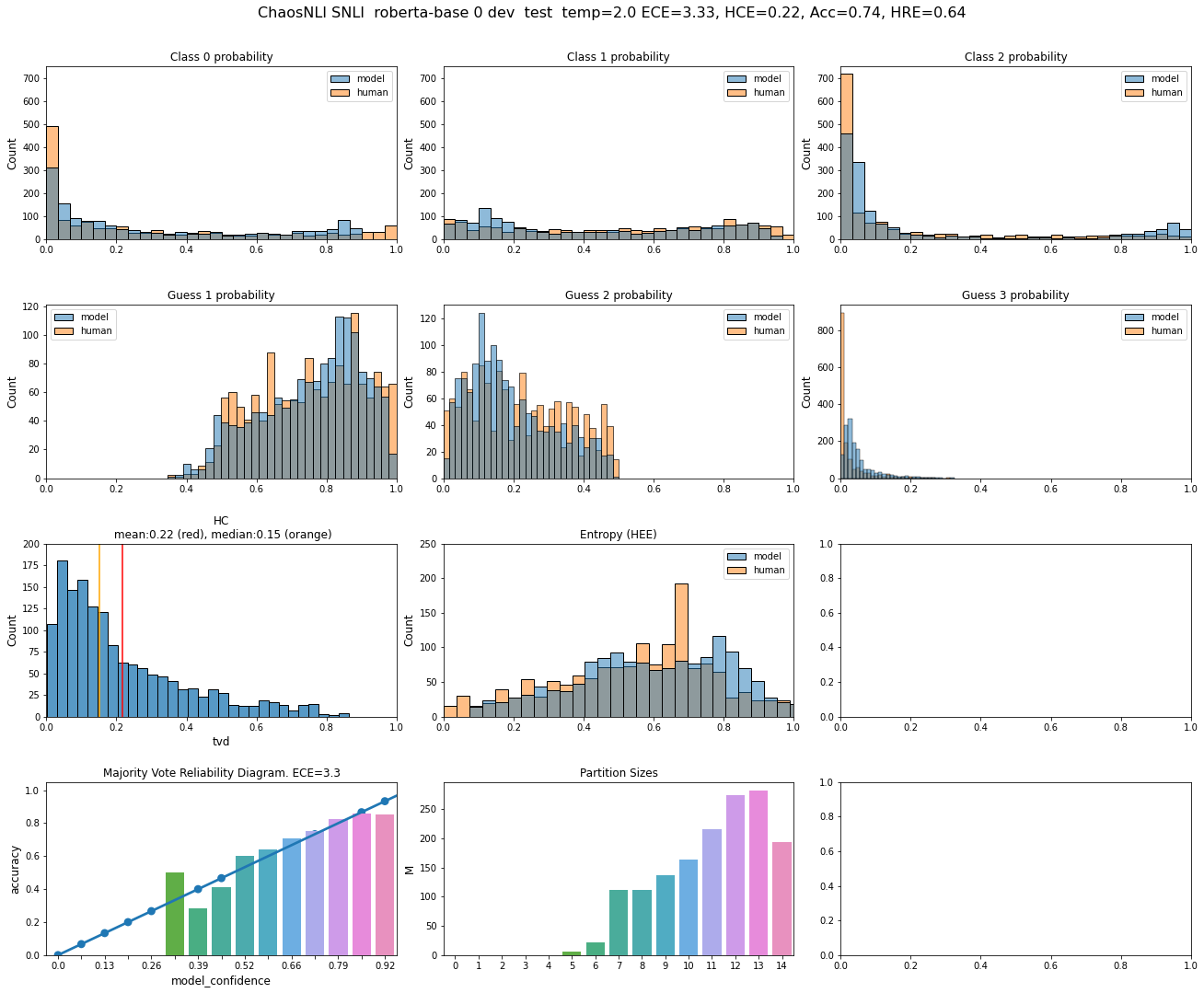}
\caption{RoBERTa-0 with oracle temperature scaling on the ChaosNLI-SNLI dev+test set. Several figures comparing human uncertainty to model uncertainty using TVD, confidence, entropy, and reliability diagrams. This figure shows the human and model distribution over the probability, entropy or TVD range. Top row shows distribution over probability magnitudes for class 0, 1 and 2, while the second row shows the distribution for first, second and third guess for the model vs human $k$th guess (i.e., the highest model probability versus the highest human probability on each instance). See Appendix \ref{ap:experiments} for more information.}
\label{figure:calibration-fig-snli-full-temp2.0}
\end{figure*}

Figure \ref{figure:calibration-fig-snli-full-temp1.0} shows a vanilla RoBERTa on ChaosNLI-SNLI. We can see that all histograms are very different between humans and model. It is clear that the model predictions are drawn from a different distribution than the human predictions. This signals bad calibration. 
Figure \ref{figure:calibration-fig-snli-full-temp2.0} shows a temperature scaled RoBERTa. According to all plots, the predictions still appear drawn from a different distribution, even though the ECE drops significantly. TS seems to transform the distributions, but they are not clearly closer to the human distribution. The entropy figure seems to indicate that TS \textit{overshoots} to the other end of the spectrum (\ie from over-confidence to under-confidence).
The TS model is unable to match the extreme left and right end of the human certainty spectrum for class 0, 1 and 2, which corroborates our intuition that the probability range is compressed. Another observation is that the human distribution rarely put much mass on the third guess (\ie the predicted class with the lowest probability). However, after TS models actually do so---which is undesirable.

\begin{figure*}[ht]
\includegraphics[width=0.65\textheight]{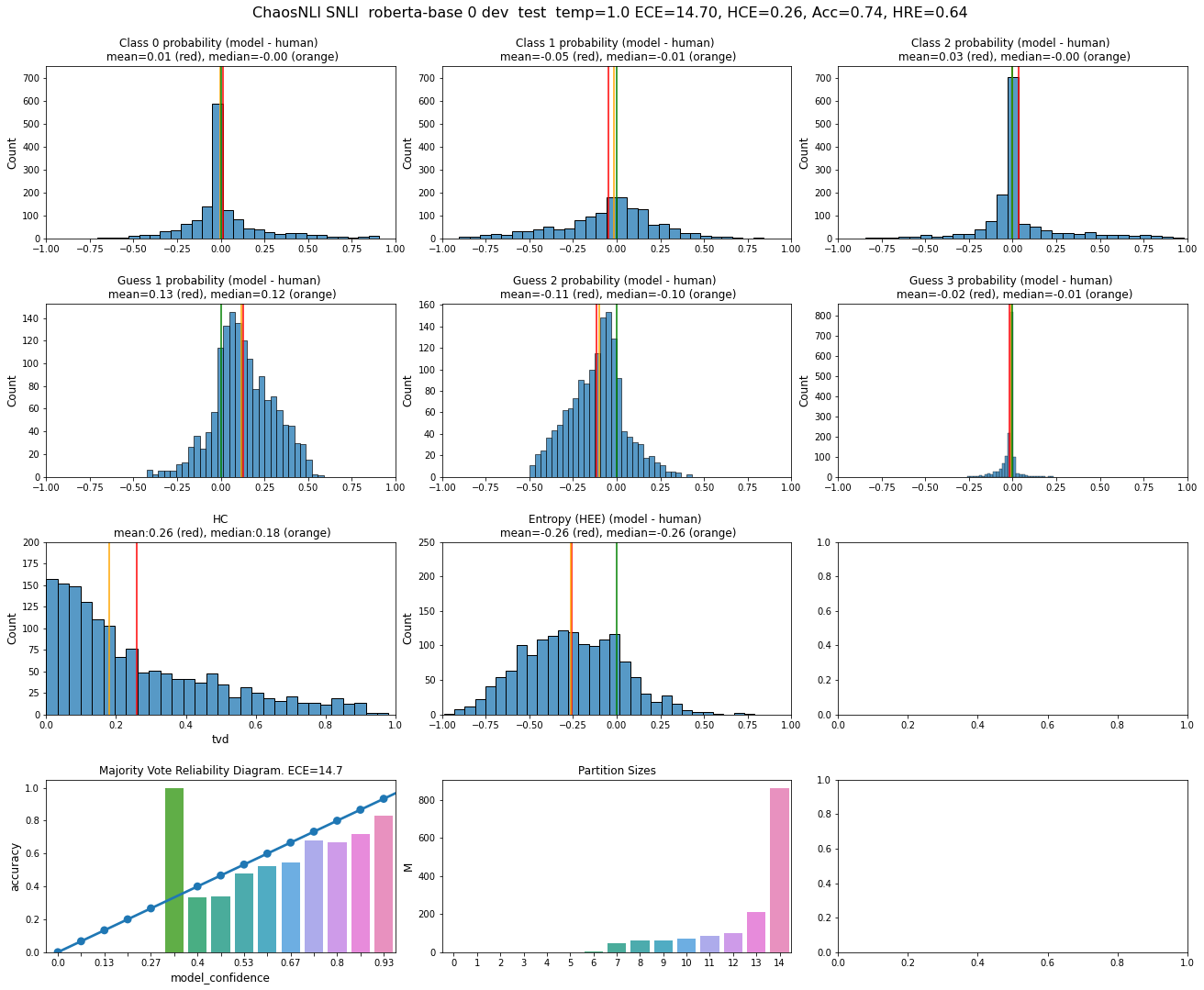}
\caption{RoBERTa-0 on the ChaosNLI-SNLI dev+test set. Several figures comparing human uncertainty to model uncertainty using TVD, confidence, entropy, and reliability diagrams. This figure shows the distribution over instance-based absolute errors between probabilities for each class (top row) or the model vs human $k$th guess (i.e., the highest model probability versus the highest human probability on each instance). See Appendix \ref{ap:experiments} for more information.}
\label{figure:calibration-fig-snli-error-temp1.0}
\end{figure*}

\begin{figure*}[ht]
\includegraphics[width=0.65\textheight]{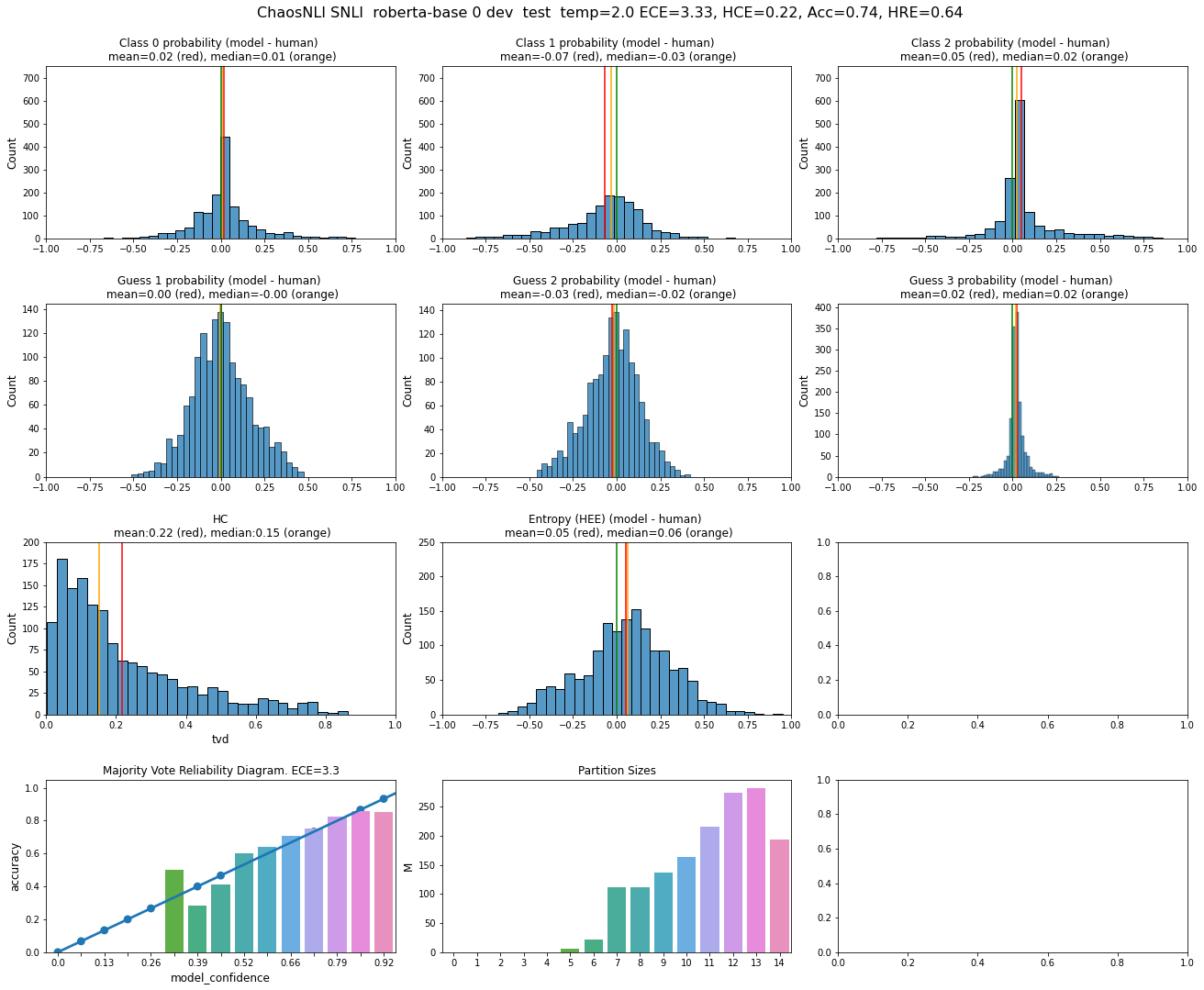}
\caption{RoBERTa-0 with oracle temperature scaling on the ChaosNLI-SNLI dev+test set. Several figures comparing human uncertainty to model uncertainty using TVD, confidence, entropy, and reliability diagrams. This figure shows the distribution over instance-based absolute errors between probabilities for each class (top row) or the model vs human $k$th guess (i.e., the highest model probability versus the highest human probability on each instance). See Appendix \ref{ap:experiments} for more information.}
\label{figure:calibration-fig-snli-error-temp2.0}
\end{figure*}

We next compare the histogram of (non absolute) instance-level errors in Figure \ref{figure:calibration-fig-snli-error-temp1.0} and \ref{figure:calibration-fig-snli-error-temp2.0}.
TS brings entropy error median and mean from -.26 to .12. The TS model therefore \textit{overshoots} also on the instance-level (recall this plot is the instance-level error, unlike the previous marginal figure we discussed) and becomes more much more uncertain than humans are. 
TS causes less class 0 predictions to have 0 error (which is bad). It also narrows the spread slightly for class 1 and increases errors for class 3.

\begin{figure*}[ht]
\includegraphics[width=0.65\textheight]{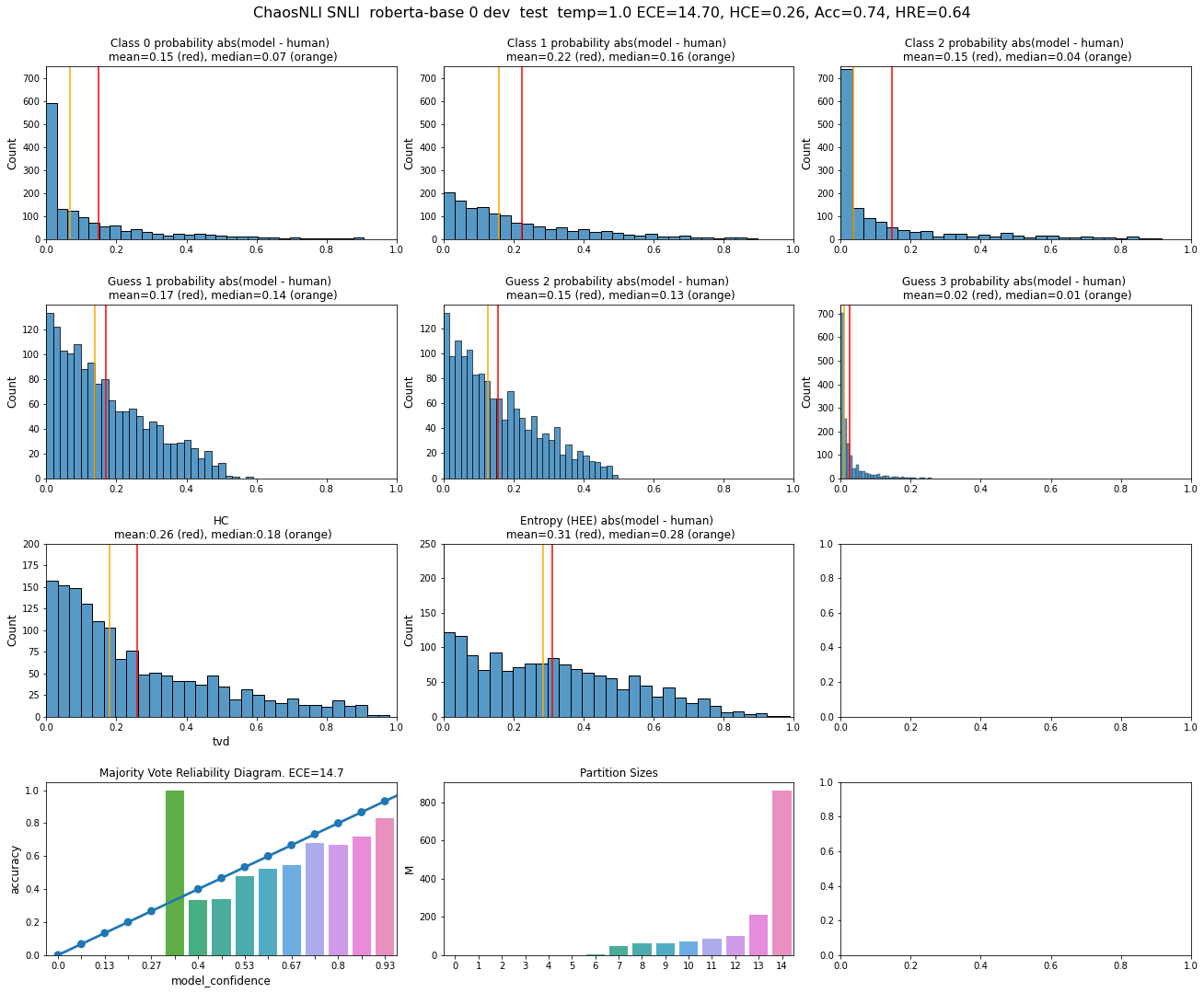}
\caption{RoBERTa-0 on the ChaosNLI-SNLI dev+test set. Several figures comparing human uncertainty to model uncertainty using TVD, confidence, entropy, and reliability diagrams. This figure shows the distribution over instance-based absolute errors between probabilities for each class (top row) or the model vs human $k$th guess (i.e., the highest model probability versus the highest human probability on each instance). See Appendix \ref{ap:experiments} for more information.}
\label{figure:calibration-fig-snli-abs_error-temp1.0}
\end{figure*}

\begin{figure*}[ht]
\includegraphics[width=0.65\textheight]{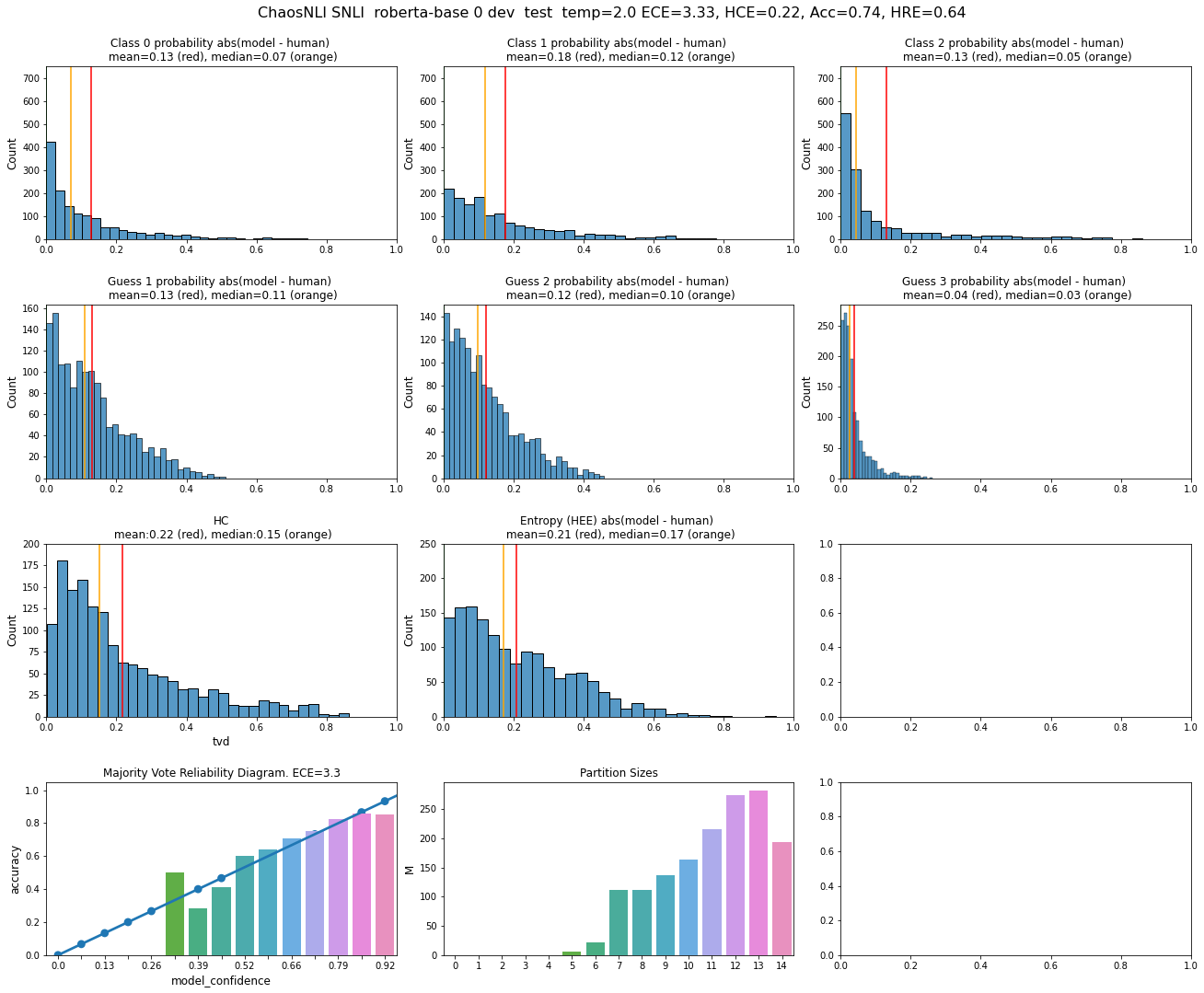}
\caption{RoBERTa-0 with oracle temperature scaling on the ChaosNLI-SNLI dev+test set. Several figures comparing human uncertainty to model uncertainty using TVD, confidence, entropy, and reliability diagrams. This figure shows the distribution over instance-based absolute errors between probabilities for each class (top row) or the model vs human $k$th guess (i.e., the highest model probability versus the highest human probability on each instance). See Appendix \ref{ap:experiments} for more information.}
\label{figure:calibration-fig-snli-abs_error-temp2.0}
\end{figure*}

We next compare the histogram of absolute errors in Figure \ref{figure:calibration-fig-snli-abs_error-temp1.0} and \ref{figure:calibration-fig-snli-abs_error-temp2.0}.
We see that TS reduces error tail of class 0, but also reduces instances with 0 error (similar trend as discussed in \S\ref{ap:experiments} about $\DistCE$). Similarly for class 2 and 3, and entropy: it seemingly reduces the mean and median by cutting off the tail. 
Finally, the human rank calibration error (RCE) shown in the title of each figure, shoes that models are not good at matching the human rank at all---as expected, much worse than matching the majority vote.

In general, it appears that the median and mean error on most metrics go down with TS, mainly by removing instances from the tail (those are extremely miscalibrated examples). However, they seem to sacrifice predictions that were well/perfectly aligned with human judgement probabilities. This illustrated by the mode of the error distributions moving towards the right (meaning more predictions with a higher error). Arguably, this is not desirable---and our metrics provide tools to expose such behaviors.

\subsection{Out of Distribution Evaluation} \label{ap:ood}
The OOD setting is interesting to evaluate uncertainty estimates, because it is especially important to have reliable uncertainty estimates for examples that are especially difficult (\eg because the classifier has not seen them during training, and might not reflect the learned distribution). In such cases, it is desirable that a classifier is more uncertain on such examples. In fact, OOD detection is often used to evaluate uncertainty estimates, next to or instead of calibration. Models are often found to be more badly miscalibrated for OOD datasets \cite{desai-durrett-2020-calibration}, which is an observation we confirm.

\begin{figure*}[ht]
\includegraphics[width=0.65\textheight]{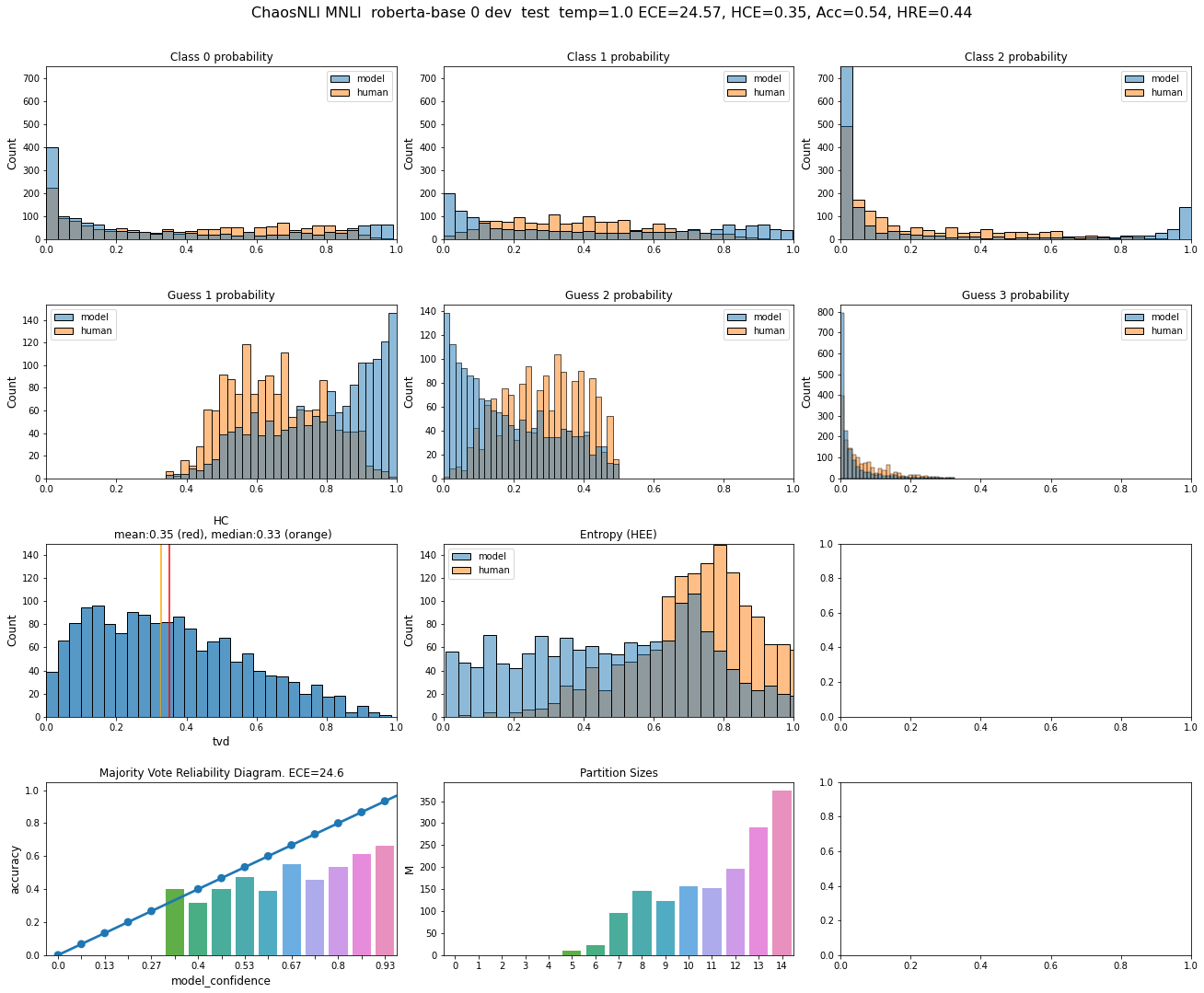}
\caption{RoBERTa-0 with (ID) temperature scaling on the OOD ChaosNLI-MNLI dev+test set. Several figures comparing human uncertainty to model uncertainty using TVD, confidence, entropy, and reliability diagrams. This figure shows the distribution over instance-based absolute errors between probabilities for each class (top row) or the model vs human $k$th guess (i.e., the highest model probability versus the highest human probability on each instance). See Appendix \ref{ap:experiments} for more information.}
\label{figure:calibration-fig-mnli-full-temp1.0}
\end{figure*}

\begin{figure*}[ht]
\includegraphics[width=0.65\textheight]{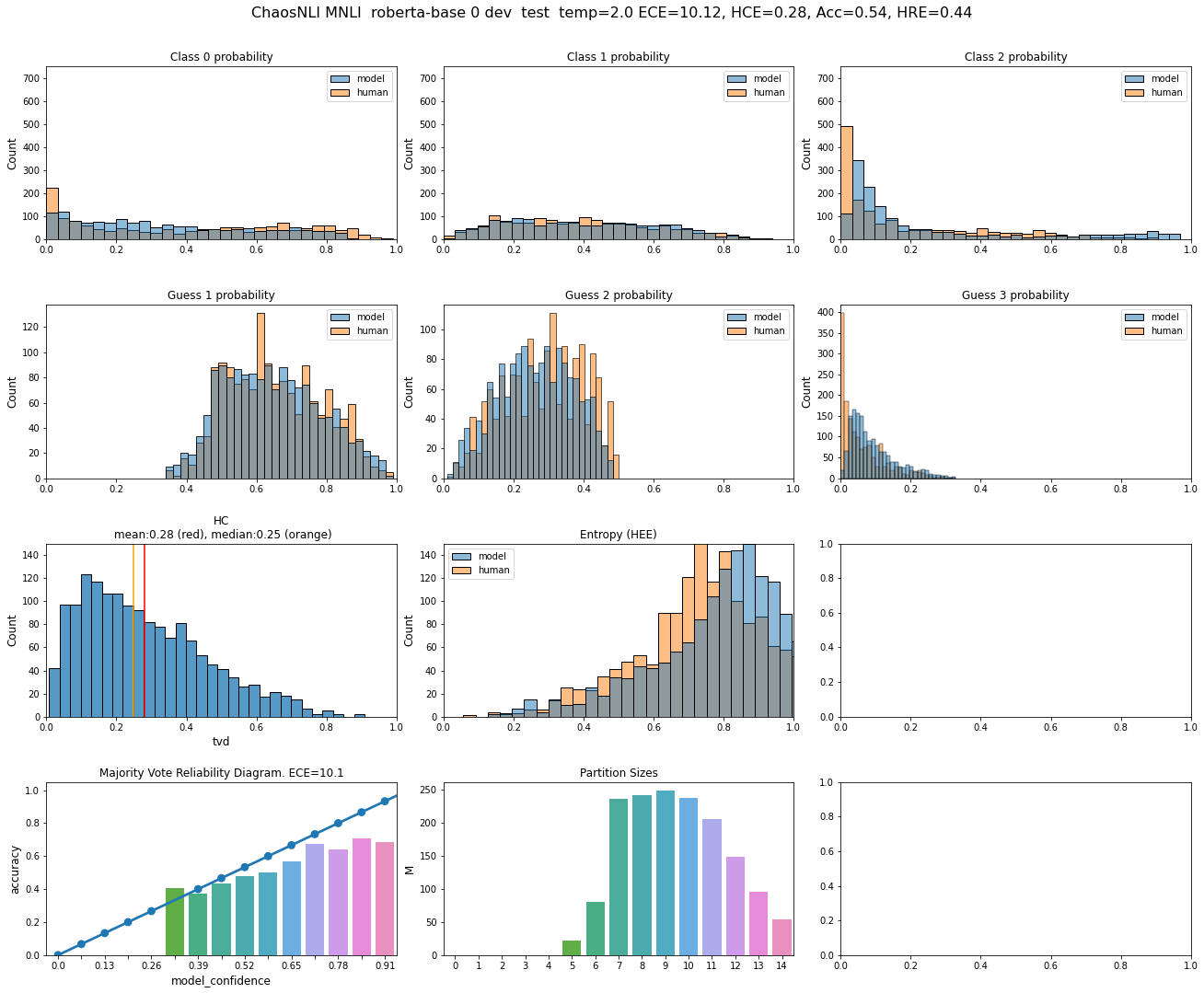}
\caption{RoBERTa-0 with (ID) temperature scaling on the OOD ChaosNLI-MNLI dev+test set. Several figures comparing human uncertainty to model uncertainty using TVD, confidence, entropy, and reliability diagrams. This figure shows the human and model distribution over the probability, entropy or TVD range. Top row shows distribution over probability magnitudes for class 0, 1 and 2, while the second row shows the distribution for first, second and third guess for the model vs human $k$th guess (i.e., the highest model probability versus the highest human probability on each instance). See Appendix \ref{ap:experiments} for more information.}
\label{figure:calibration-fig-mnli-full-temp2.0}
\end{figure*}

\begin{figure*}[ht]
\includegraphics[width=0.65\textheight]{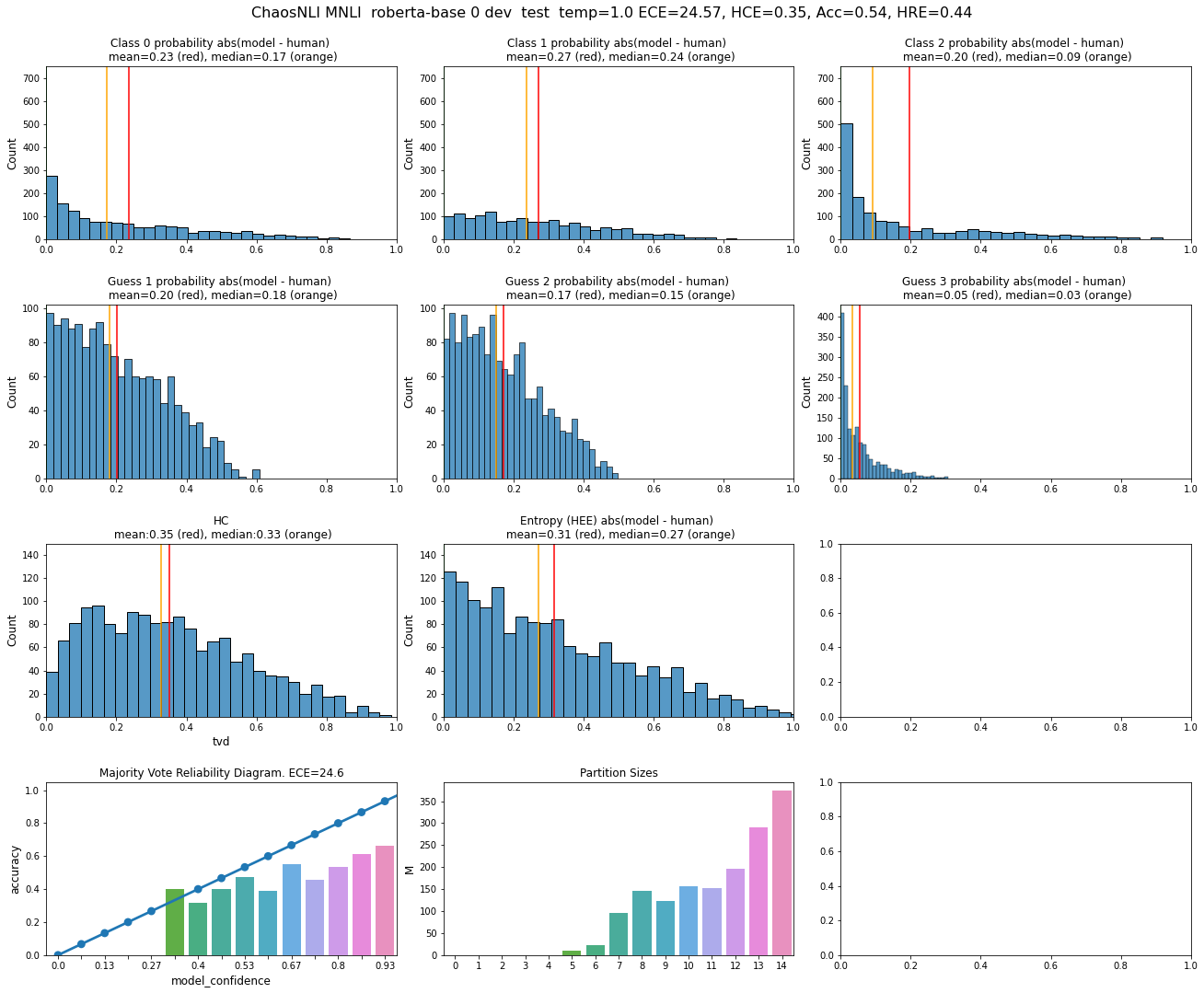}
\caption{RoBERTa-0 with (ID) temperature scaling on the OOD ChaosNLI-MNLI dev+test set. Several figures comparing human uncertainty to model uncertainty using TVD, confidence, entropy, and reliability diagrams. This figure shows the distribution over instance-based absolute errors between probabilities for each class (top row) or the model vs human $k$th guess (i.e., the highest model probability versus the highest human probability on each instance). See Appendix \ref{ap:experiments} for more information.}
\label{figure:calibration-fig-mnli-abs_error-temp1.0}
\end{figure*}

\begin{figure*}[ht]
\includegraphics[width=0.65\textheight]{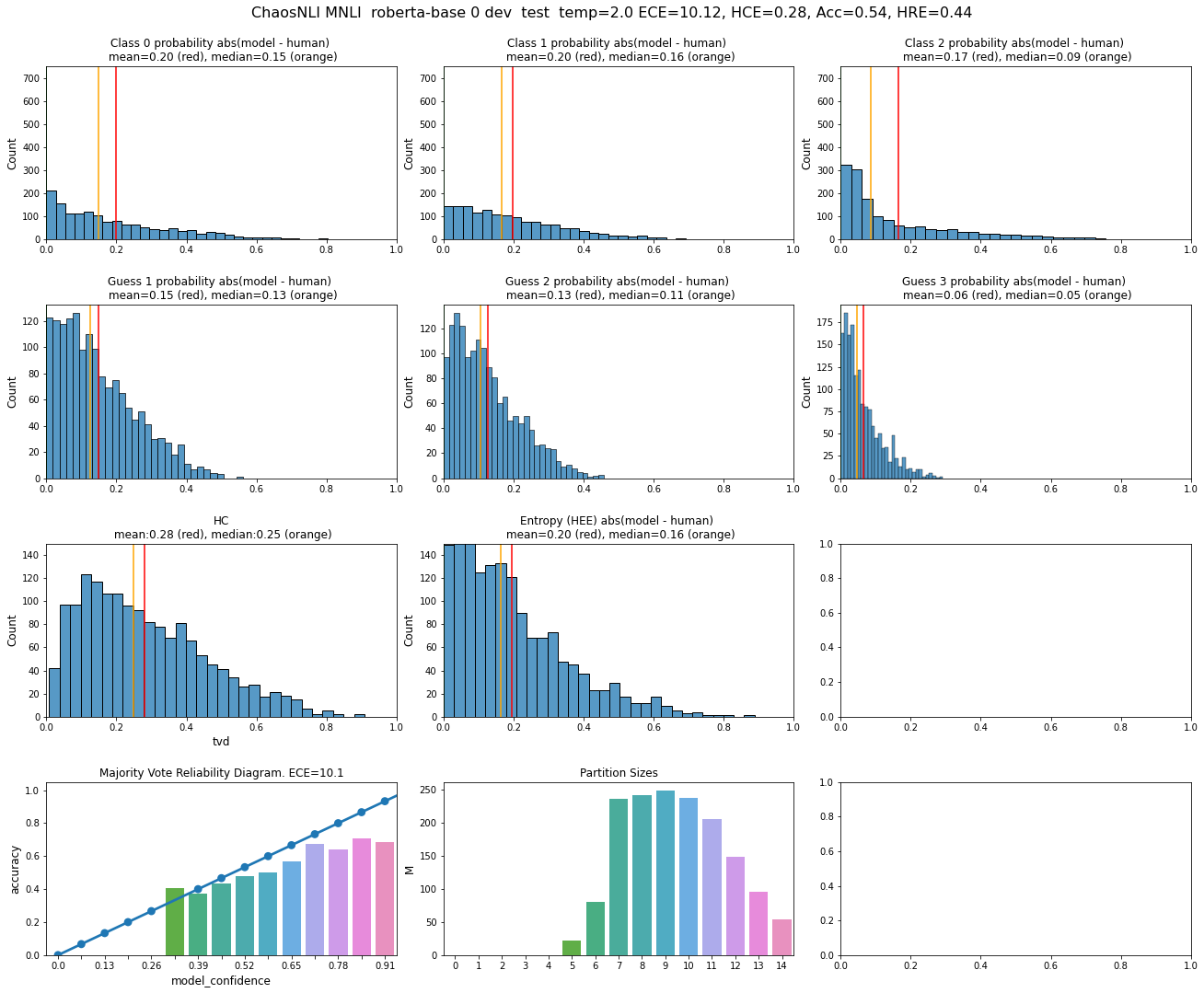}
\caption{RoBERTa-0 with oracle temperature scaling on the ChaosNLI-SNLI dev+test set. Several figures comparing human uncertainty to model uncertainty using TVD, confidence, entropy, and reliability diagrams. This figure shows the distribution over instance-based absolute errors between probabilities for each class (top row) or the model vs human $k$th guess (i.e., the highest model probability versus the highest human probability on each instance). See Appendix \ref{ap:experiments} for more information.}
\label{figure:calibration-fig-mnli-abs_error-temp2.0}
\end{figure*}

\begin{figure*}[ht]
\includegraphics[width=0.65\textheight]{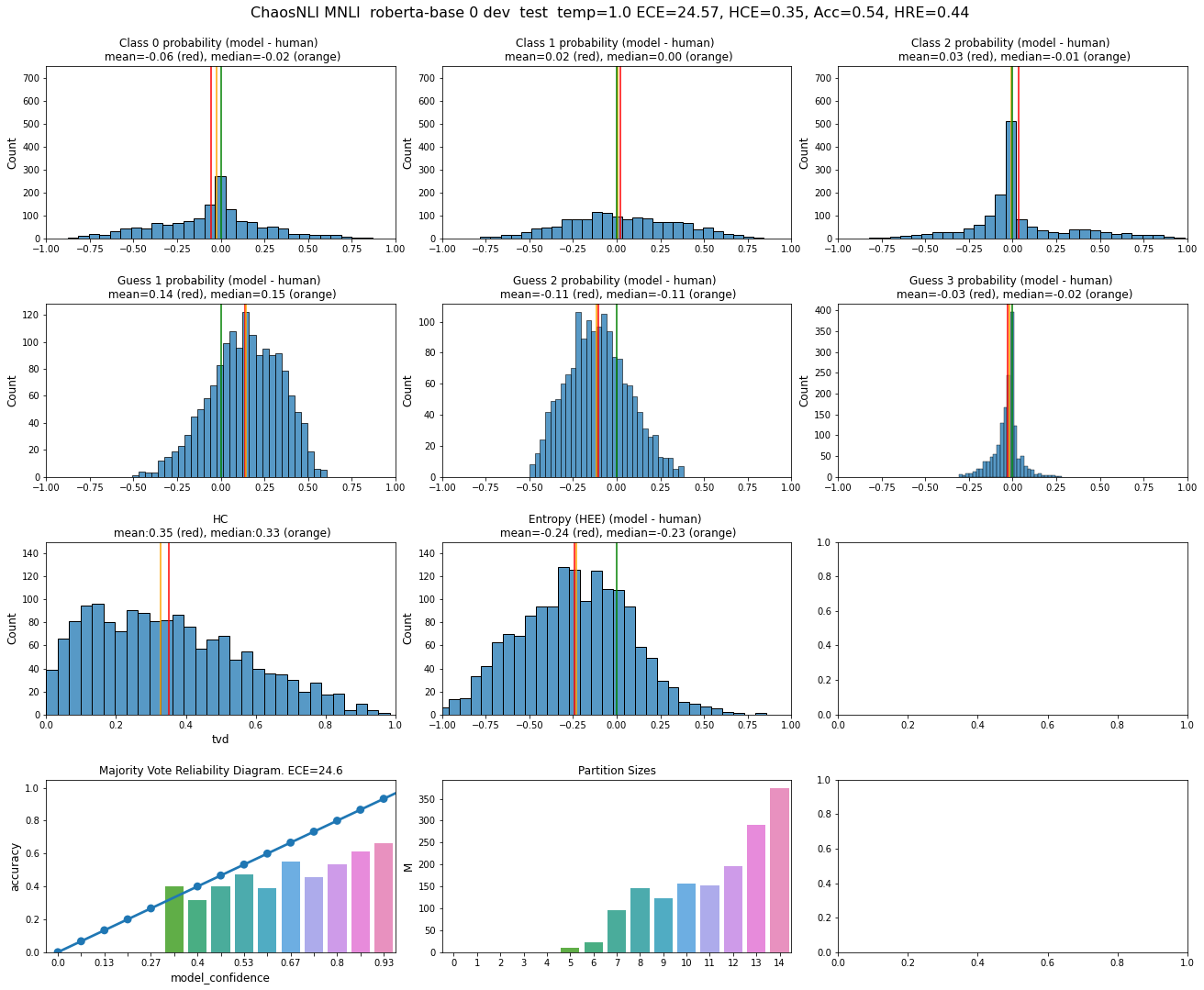}
\caption{RoBERTa-0 with (ID) temperature scaling on the OOD ChaosNLI-MNLI dev+test set. Several figures comparing human uncertainty to model uncertainty using TVD, confidence, entropy, and reliability diagrams. This figure shows the distribution over instance-based absolute errors between probabilities for each class (top row) or the model vs human $k$th guess (i.e., the highest model probability versus the highest human probability on each instance). See Appendix \ref{ap:experiments} for more information.}
\label{figure:calibration-fig-mnli-error-temp1.0}
\end{figure*}

\begin{figure*}[ht]
\includegraphics[width=0.65\textheight]{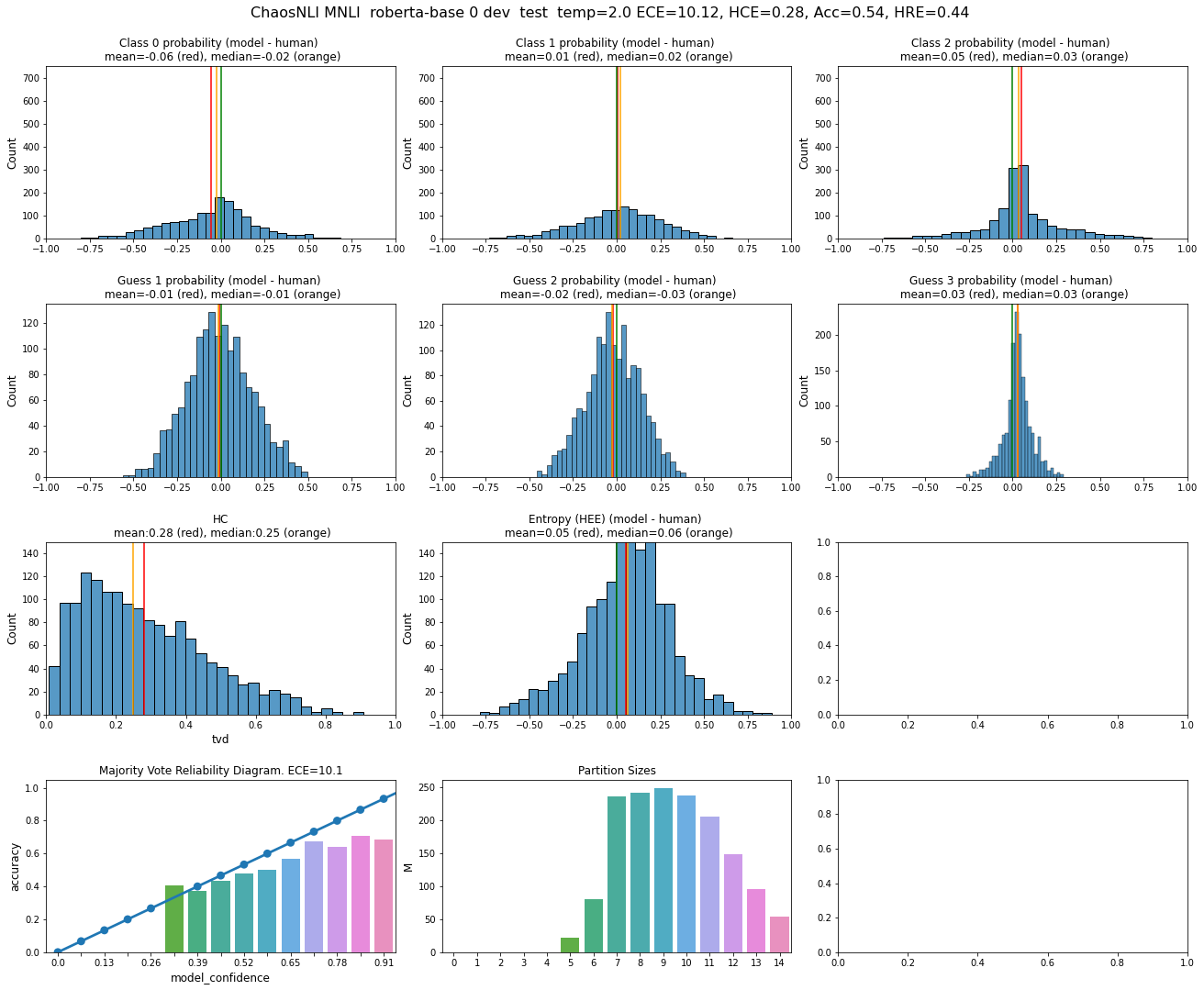}
\caption{RoBERTa-0 with (ID) temperature scaling on the OOD ChaosNLI-MNLI dev+test set. Several figures comparing human uncertainty to model uncertainty using TVD, confidence, entropy, and reliability diagrams. This figure shows the distribution over instance-based absolute errors between probabilities for each class (top row) or the model vs human $k$th guess (i.e., the highest model probability versus the highest human probability on each instance). See Appendix \ref{ap:experiments} for more information.}
\label{figure:calibration-fig-mnli-error-temp2.0}
\end{figure*}

We observe similar trends as in the in-distribution analysis. The marginal distributions from Figure \ref{figure:calibration-fig-mnli-full-temp1.0} to Figure \ref{figure:calibration-fig-mnli-full-temp2.0} seem to match the human marginal slightly better than on the in-distribution dataset. However, inspecting the error distributions in Figure \ref{figure:calibration-fig-mnli-error-temp1.0}, \ref{figure:calibration-fig-mnli-error-temp2.0}, 
\ref{figure:calibration-fig-mnli-abs_error-temp1.0} and \ref{figure:calibration-fig-mnli-abs_error-temp2.0}, we see that the distributions are transformed somewhat, but we do not believe that to be evidence that TS is a good method to improve calibration. The instances are still obviously drawn from a different distribution.

\section{Temperature Scaling} \label{ap:ts}
Temperature scaling is a simple method that uses a single temperature parameter $t$ to scale the output logits of a classifier \cite{guo-on-calibration-2017}. The standard way to choose a temperature is to perform a search on a range of possible values for $t$ on a development set. However, sometimes, the temperature is tuned directly on the test set. This is commonly referred to as the \textit{oracle temperature}. Indeed, we use this method to obtain our temperature, because we consider it an (unrealistic) upper bound on what TS can do. For OOD evaluation, we use the temperature tuned on the ID evaluation. In our experiments in \S\ref{section:experiments}, we found a temperature of $2.0$ to result in the lowest ECE.

\section{Classwise-ECE} \label{ap:classwise-ece}
Classwise-ECE~\cite{nixon_measuring-2019} is based on the notion of classwise-calibration~\cite{vaicenavicius2019evaluating, kull2019beyond}. The main difference with ECE---based on the notion of confidence calibration (\S\ref{section:background})---is that it removes the dependency on a decision rule on top of the classifier, and computes the calibration separately for each class. Classwise-ECE, then, is the average calibration error over classes:

\begin{align}
    \label{eq:classwise-ece}
    \frac{1}{K}\sum_{k=1}^K\sum_{m=1}^M\frac{|B_{mk}|}{N}|\mathrm{acc}(m,k)-\mathrm{conf}(m,k))|
\end{align}

Table \ref{ap:classwise-ece} shows a similar trend for classwise-ECE as for ECE: the oracle classifier is severely miscalibrated. This confirms our claims that the general notion of calibration is not suited for data on which humans inherently disagree about a class---and is not restricted to \citet{guo-on-calibration-2017} notion of confidence calibration.
\begin{table}[ht]
\centering
\resizebox{0.7\columnwidth}{!}{
\begin{tabular}{@{}cccc@{}}
\toprule
    & RoBERTa & RoBERTa-TS & Oracle \\ \midrule
    Classwise-ECE & 10      & 5          & 16     \\
\bottomrule
\end{tabular}
}
\caption{\label{tab:classwise-ece} Classwise-ECE on ChaosNLI-SNLI in \%.} 
\vspace{-\baselineskip}
\end{table}

\section{Total Variation Distance} \label{ap:TVD}

The total variation distance $\operatorname{TVD}(\mathbf q, \mathbf p) $ between two Categorical distributions with parameters $\mathbf q, \mathbf p \in \Delta_{C-1}$ is defined as: 
\begin{subequations}
\begin{align}
    &\max_{\mathcal A \in \mathcal P([C])}~|\mathbb Q(\mathcal A)-\mathbb P(\mathcal A)| \label{eq:TVD-def} \\
    \intertext{which can also be expressed as}
    &=\frac{1}{2}\underbrace{\sum_{c=1}^C |q_c - p_c|}_{= \norm{\mathbf q - \mathbf p}_1} ~. \label{eq:TVD-L1}
\end{align}
\end{subequations}
where $[C]$ is the sample space,  $\mathcal P([C])$ is the event space (for generality, we use the powerset of $[C]$, the set of all subsets of outcomes in the sample space), $\mathcal A \in \mathcal P([C])$ is any event in the event space, $\mathbb Q$ and $\mathbb P$ are the probability measures prescribed by each of the Categorical distributions (\ie 
$\mathbb Q(\mathcal A) = \sum_{c \in \mathcal A} q_a$ and $\mathbb P(\mathcal A) = \sum_{c \in \mathcal A} p_a$). For a complete technical results with definitions, proofs, and various properties see 
\citet[Chapter 5]{devroye2001combinatorial}.

\paragraph{Properties.} TVD is defined for any two probability vectors, whether dense or sparse. It is a metric (hence symmetric and minimised only for identical distributions) and bounded:
\begin{subequations}
\begin{align}
    &\operatorname{TVD}(\mathbf q, \mathbf p) = \operatorname{TVD}(\mathbf p, \mathbf q) \\
    &\operatorname{TVD}(\mathbf q, \mathbf p)=0 \quad \text{ iff  } \mathbf q = \mathbf p \\
    &\operatorname{TVD}(\mathbf q, \mathbf p) \in [0, 1]~.
\end{align}
\end{subequations}

\paragraph{Interpretations.} The identity in Eq(\ref{eq:TVD-L1}), which expresses TVD in terms of the L1 norm, gives us a rather practical (linear-time) algorithm to compute it by summing half the absolute difference in probability for the outcomes in the sample space $[C]$ of the random variable. This means that TVD is expressed in units of absolute difference in probability. 
That definition, Eq(\ref{eq:TVD-def}),  also helps interpretation, it shows that TVD quantifies the maximum discrepancy in probability between the two measures over their entire event spaces.


\end{document}